%% file: acl_latex.tex
\documentclass[11pt]{article}

\usepackage[preprint]{acl}

\usepackage{times}
\usepackage{latexsym}

\usepackage[T1]{fontenc}

\usepackage[utf8]{inputenc}

\usepackage{microtype}

\usepackage{inconsolata}

\usepackage{graphicx}
\usepackage{xspace}
\usepackage{enumitem}

\usepackage{xcolor}

\newcommand{\task}{\textsc{UMUI}\xspace}
\newcommand{\method}{\textsc{CLUE}\xspace}
\newcommand{\wikivideo}{\textsc{WikiVideo}\xspace}
\newcommand{\multivent}{\textsc{MultiVENT2.0}\xspace}
\newcommand{\unli}{\textsc{UNLI}\xspace}
\newcommand{\clotho}{\textsc{Clotho}\xspace}

\title{Unified Multimodal Uncertain Inference}

\author{Dengjia Zhang\textsuperscript{\rm 1}  
\quad Alexander Martin\textsuperscript{\rm 1} 
\quad William Jurayj \textsuperscript{\rm 1}\\
\quad \textbf{Kenton Murray}\textsuperscript{\rm 1,2}
\quad \textbf{Benjamin Van Durme}\textsuperscript{\rm 1,2}
\quad \textbf{Reno Kriz}\textsuperscript{\rm 1,2}
\\
  \textsuperscript{1}Johns Hopkins University\quad \textsuperscript{2}Human Language Technology Center of Excellence\quad \\
  \texttt{\small{\{dzhang98, amart233, rkriz\}@jhu.edu}}}

\begin{document}
\maketitle
\begin{abstract}

\input{sections/00-abstract}

\end{abstract}


\section{Introduction}

\input{sections/10-intro}

\section{\task}
\input{sections/30-data}

\section{\method}
\input{sections/40-method}

\section{Experiments}
\input{sections/50-results}

\section{Conclusion}
\input{sections/70-conclusion}






\bibliography{custom} 

\appendix

\section{Extended Related Work}
\label{append:related}
\input{appendix/related}

\section{Annotation Protocol}
\label{append:annotation}
\input{appendix/annotation}

\section{Teacher Model Selection}
\label{append:synth_model}
\input{appendix/synth_model}

\section{Training Details}
\label{append:training}
\input{appendix/training}

\section{Temperature Scaling Analysis}
We applied temperature scaling to baseline models to evaluate post-hoc calibration on the UMUI task. Empirical results in \autoref{tab:binary_temp_scaling} show that temperature scaling yields minimal change in Mean Squared Error (MSE) compared to unscaled baselines. 

While these optimized baselines achieve MSE scores comparable to our 3B-parameter CLUE models, the lack of improvement from scaling suggests that global logit adjustment is insufficient for complex multimodal reasoning. Unlike these baselines, CLUE internalizes calibration through an end-to-end latent distribution , providing a more robust framework for fine-grained uncertainty without requiring post-training parameter tuning.

\begin{table}[]
    \centering
    \setlength{\tabcolsep}{3.5pt}
    \begin{tabular}{lc|cccc}
    \toprule
        \textbf{Model} & \textbf{P} & \textbf{Clot} & \textbf{WV-V} & \textbf{WV-A} & \textbf{WV-AV} \\
    \midrule
        AF & 7B & 49.3 & $-$ & 20.3 & $-$ \\
        Q2-A & 7B & 49.8 & $-$ & 34.7 & $-$ \\
    \midrule
        Q2.5-VL & 32B & $-$ & 24.0 & $-$ & $-$ \\
        Q3-VL & 32B & $-$ & \textbf{80.4}& $-$ & $-$ \\
    \midrule
        Q2.5-O & 7B & 50.0 & 46.2 & 66.1 & 43.4 \\
    \midrule
        \method-T & 3B & \textbf{97.5} & 56.3 & \textbf{71.5} & 54.3 \\
        \method-D & 3B & \underline{95.8} & \underline{74.6} & \underline{70.2} & \textbf{70.1}\\
    \bottomrule
    \end{tabular}
    \caption{Accuracy of binary judgments with temperature scaling. P: Parameters. AF: Audio Flamingo. Q: Qwen, A-Audio VL-Vision Language, O-Omni. Clot: Clotho, WV: \wikivideo, V-Vision only, A-Audio only, AV-Audio-Visual.}
    \label{tab:binary_temp_scaling}
    \vspace{-1em}
\end{table}

\section{Additional Results}
\label{app:additional results}
\paragraph{Calibration Metrics}
\autoref{tab:calibration_metrics} reports full calibration metrics (MSE, NLL, ECE) with 95\% bootstrap confidence intervals (2,000 resamples).

\paragraph{Reliability Diagrams}
Figure~\ref{fig:calibration} compares reliability diagrams of \method-D and the strongest zero-shot baseline (Qwen3-VL-32B) on UNLI and WV-V. On WV-V, \method-D produces a largely monotonic curve covering all confidence bins, whereas the baseline's confidence is weakly aligned with human probability: its lowest-confidence bin corresponds to a mean
human probability of 0.6, and several bins are empty. On UNLI, the baseline's lower ECE partly reflects its discrete token outputs concentrating on few confidence values, which ECE does not penalize; \method-D covers the full probability range and achieves lower MSE and NLL (Table~\ref{tab:calibration_metrics}).

\paragraph{Component Ablations}
\autoref{tab:ablation} ablates each \method{} component with the same backbone, data, and LoRA configuration. All components contribute: regression collapses to near chance on \clotho{}, and removing teacher calibration, modality batching, or the premise degrades all settings.

\paragraph{Parameter Sensitivity}
\autoref{tab:sensitivity} varies $N$ and $\sigma$. Deviation
in either direction from the default ($N{=}100$, $\sigma{=}0.05$)
degrades performance, most visibly on audiovisual.

\input{appendix/additional_results}

\section{Cross-Family Result}
\label{app:gemma}
\input{appendix/cross-family}

\section{Threshold Analysis}
\label{append:threshold}

\input{appendix/threshold}

\section{Prompts}
\label{append:prompts}
\input{appendix/prompt}

\section{Acknowledgment of AI}
AI assistants were used in this project for coding and to edit text for fluency and typos.

\begin{figure*}
    \centering
    \includegraphics[width=\linewidth]{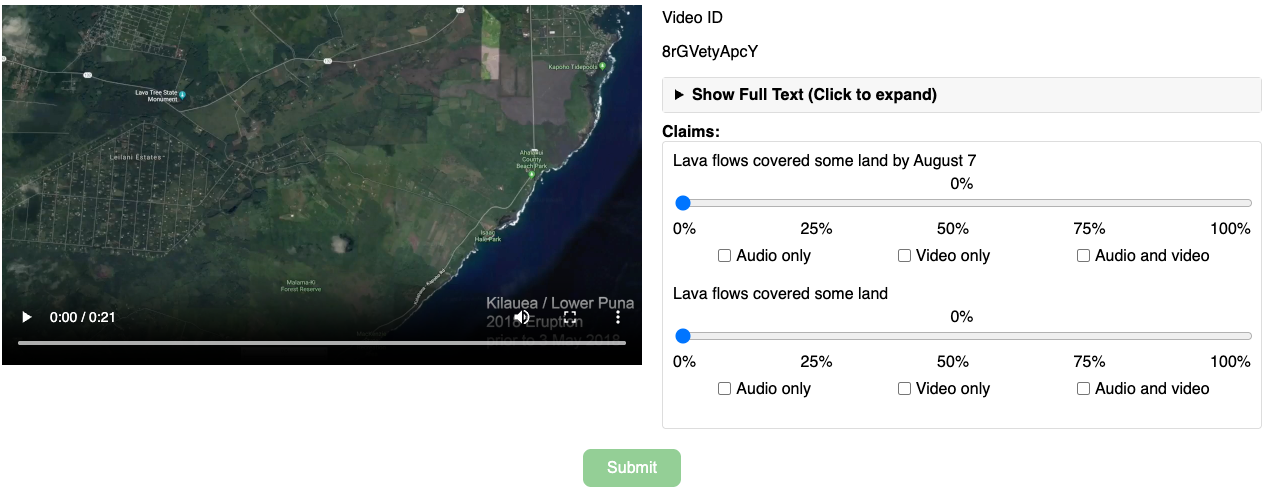}
    \caption{Annotation Protocol. Annotators are given a video (left) and set of claims (right) and can annotate the probability the claim is true with the supporting modalities.}
    \label{fig:annotation_protocol}
\end{figure*}

\begin{figure*}
    \centering
    \includegraphics[width=\linewidth]{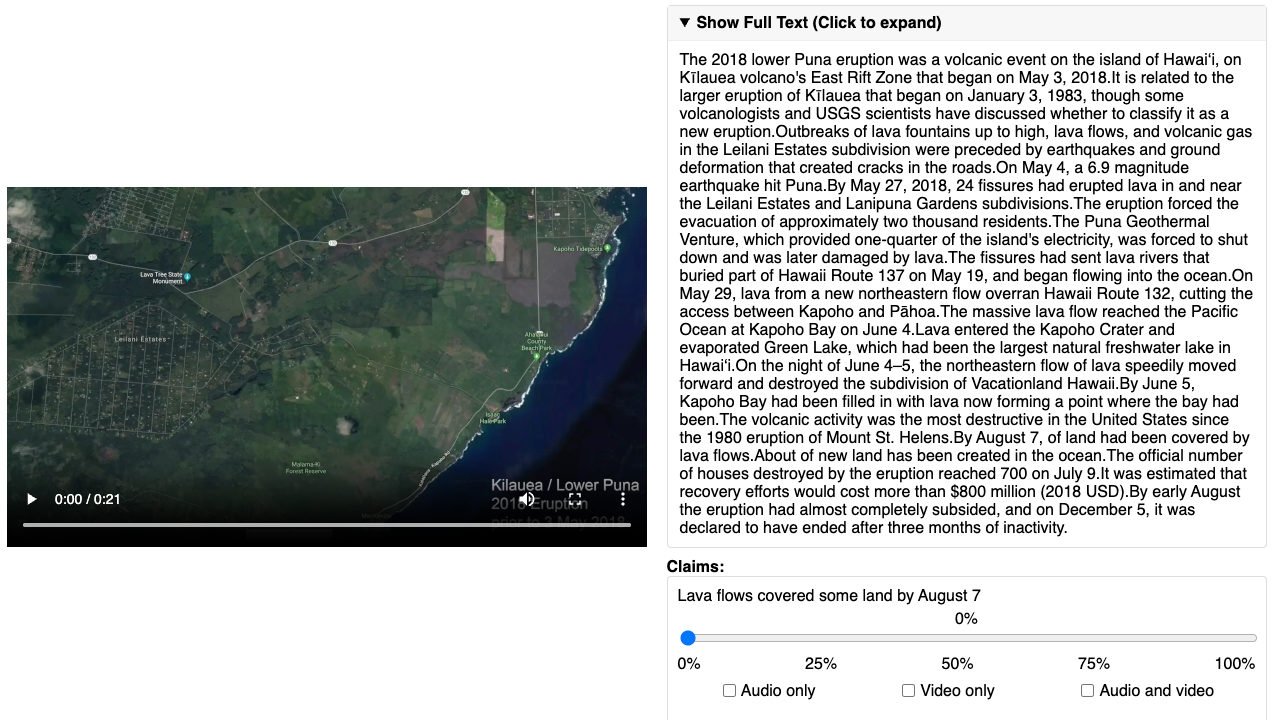}
    \caption{Annotation Protocol. Annotators can expand the annotation view to see the context from which the claim is taken to decontextualize and disambiguate any part of the claim.}
    \label{fig:annotation_protocol_context}
\end{figure*}
\input{figures/annotation_instructions}

\end{document}

%% file: sections/00-abstract.tex
We introduce Unified Multimodal Uncertain Inference (\task), a multimodal inference task spanning text, audio, and video, where models must produce calibrated probability estimates of hypotheses conditioned on a premise in any modality or combination. While uncertain inference has been explored in text, extension to other modalities has been limited to single-modality binary entailment judgments, leaving no framework for fine-grained probabilistic reasoning in or across other modalities. To address this, we curate a human-annotated evaluation set with scalar probability judgments across audio, visual, and audiovisual settings, and additionally evaluate on existing text and audio benchmarks. We introduce \method (\textbf{C}alibrated \textbf{L}atent \textbf{U}ncertainty \textbf{E}stimation), which combines self-consistent teacher calibration and distribution-based confidence probing to produce calibrated predictions. We demonstrate that our 3B-parameter model achieves equivalent or stronger performance than zero-shot baselines up to  32B parameters across all modalities.\footnote{Data, code, and models released upon publication}

%% file: sections/10-intro.tex
\begin{figure}[!t]
    \centering
    \includegraphics[width=1\linewidth]{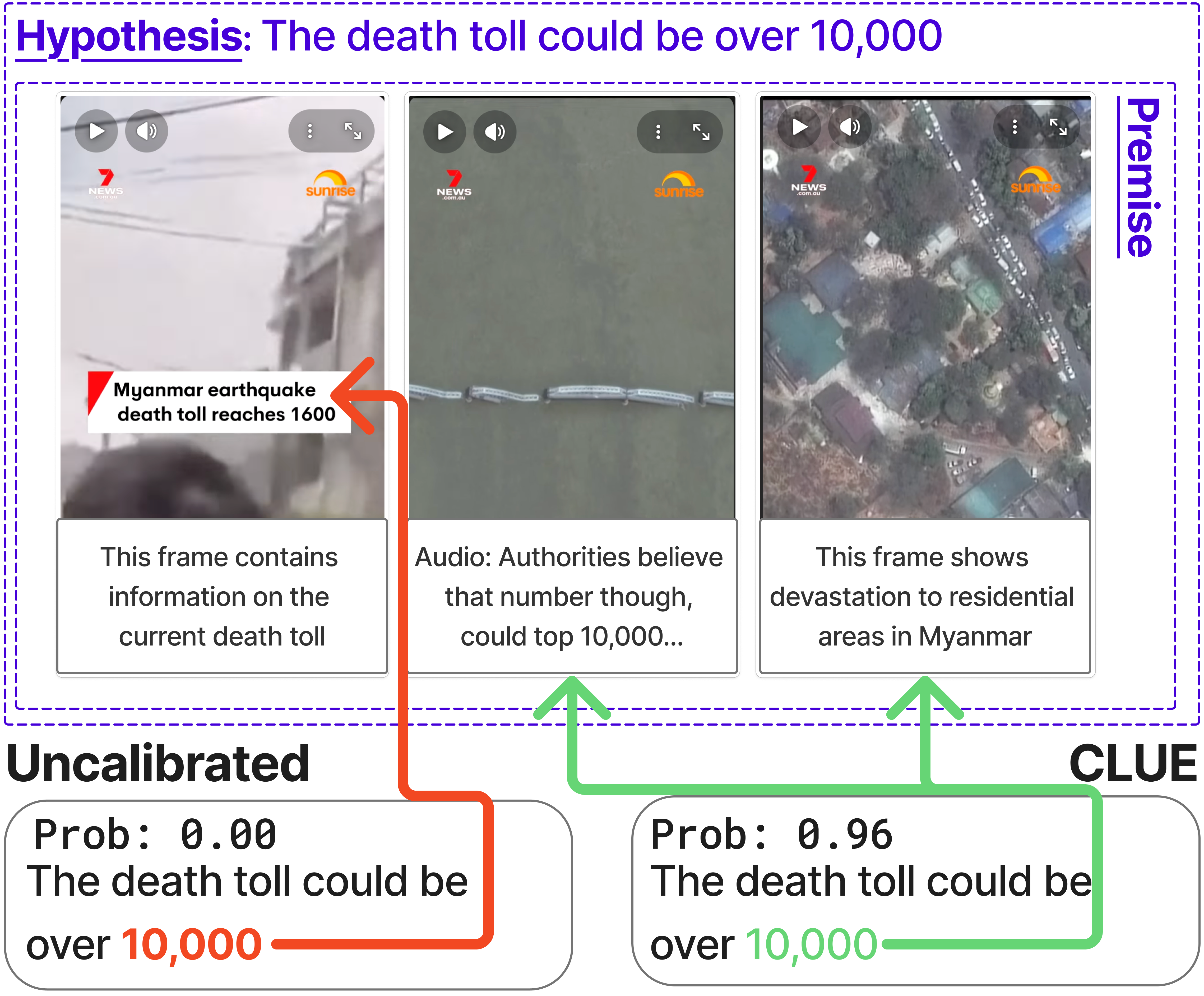}
    \caption{An example UMUI instance. An uncalibrated baseline relies solely on OCR text while ignoring supporting audio and visual evidence, whereas CLUE integrates clues across modalities to produce a calibrated estimate of the hypothesis.}
    \label{fig:teaser}
    \vspace{-2em}
\end{figure}

Human reasoning is fundamentally probabilistic, operating over degrees of belief rather than binary judgments \cite{Oaksford2001ThePA, human_cognition, tenebaum_prob_reasoning}. 
When viewing social media footage in the aftermath of an earthquake, we integrate the extent of visible structural collapse, the sounds of sirens and car alarms, and bystanders' reactions to form a graded estimate of severity -- not a binary classification, but a calibrated belief synthesized across vision and audio. More broadly, this probabilistic reasoning operates fluidly across modalities, with beliefs continuously updated the more we read, watch, and listen. As multimodal AI systems are increasingly deployed in high-stakes domains, their inability to express calibrated uncertainty over multimodal evidence limits their trustworthiness and utility (\autoref{fig:teaser}).

Uncertain Natural Language Inference (\unli) addresses this gap in text, replacing categorical entailment labels with scalar probability estimates that better reflect human judgment \cite{chen-etal-2020-uncertain}. However, uncertainty-aware inference remains almost exclusively textual \cite{geng2024surveyconfidenceestimationcalibration, xiong2024can, wang2025always}. Extensions to other modalities remain limited: in vision, calibrated uncertainty has been explored for event classification \cite{sanders2022ambiguous}, while most inference tasks in audio and video have been formulated only as binary entailment judgments \cite{xie2019visualentailmentnoveltask, liu2020violinlargescaledatasetvideoandlanguage, deshmukh2024audioentailmentassessingdeductive, martin2025mirage}. As a result, no existing framework supports general probabilistic inference in or across non-textual modalities.

\input{figures/main_figure}
We address these limitations by introducing Unified Multimodal Uncertain Inference (\task), a task for calibrated probabilistic inference spanning text, audio, video, and their combinations. To support this task, we collect human scalar probability judgments for audio-visual premise-hypothesis pairs. We find that existing multimodal language models are poorly calibrated for this task, motivating \method (\textbf{C}alibrated \textbf{L}atent \textbf{U}ncertainty \textbf{E}stimation), a training approach that combines self-consistent teacher calibration, distribution-based confidence probing, and modality-specific batching to produce well-calibrated predictions from a single multimodal model.

Our contributions are: (1)~We formalize \task, the task of producing calibrated probability estimates of hypotheses conditioned on a premise in any modality or combination; (2)~We collect human scalar probability judgments for audio-visual premise-hypothesis pairs, extending existing binary annotations to fine-grained uncertainty labels; and (3)~We introduce \method and demonstrate that our 3B-parameter model achieves better performance than zero-shot baselines up to  32B parameters across text, audio, visual, and audiovisual settings.

%% file: figures/main_figure.tex
\begin{figure*}
\vspace{-1em}
    \centering
    \includegraphics[width=\linewidth]{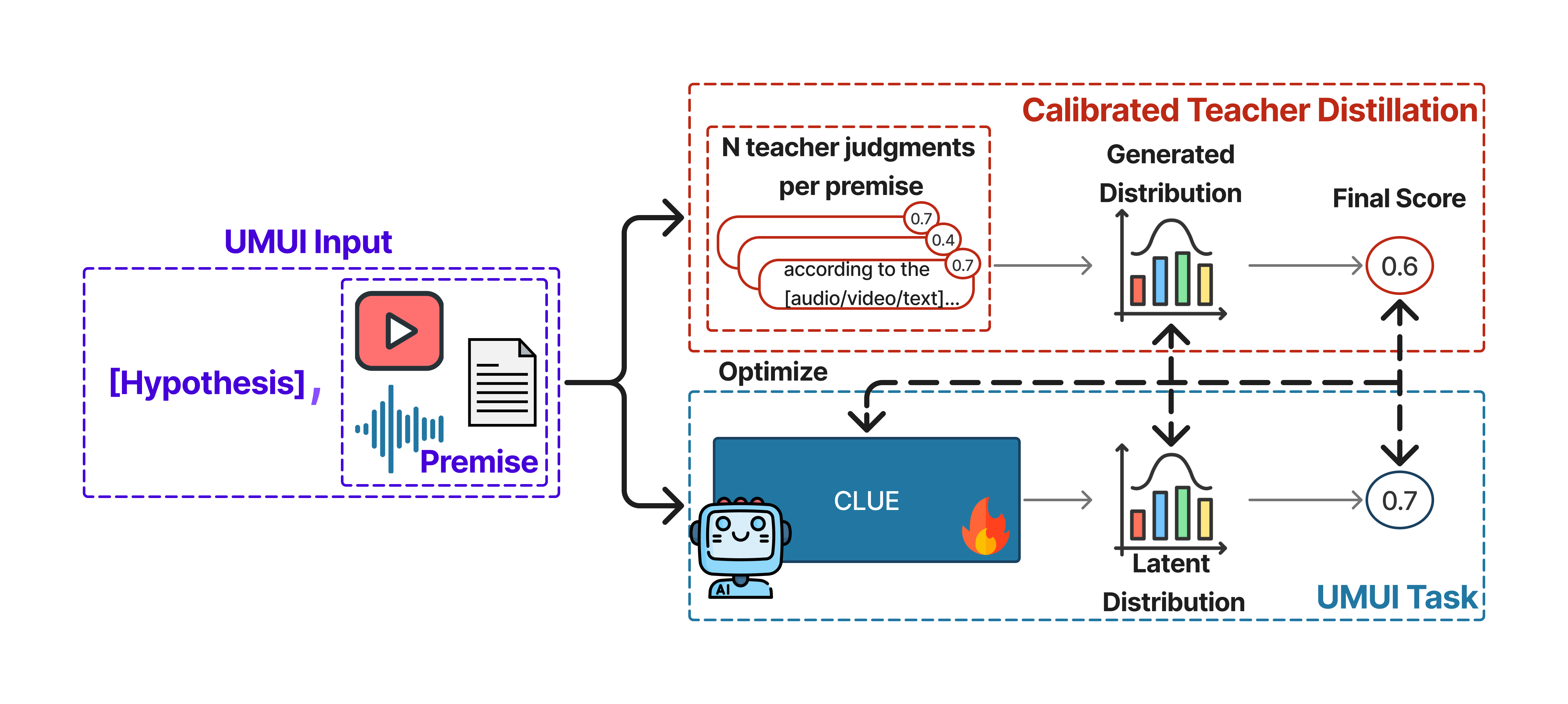}
    \vspace{-4em}
    \caption{Overview of \task and \method. \task takes a hypothesis and multimodal premise (audio, video, and/or text) and produces a calibrated probability estimate of the hypothesis. \textbf{Top right:} To generate training labels, we produce N independent teacher judgments per premise-hypothesis pair and aggregate them into calibrated distillation scores. \textbf{Bottom right:} \method takes a hypothesis-premise pair in any modality and predicts a latent distribution over confidence tokens, which is aggregated into a scalar probability estimate.}
    \label{fig:method}
    \vspace{-1.5em}
\end{figure*}

%% file: sections/30-data.tex
\paragraph{Task Definition}
The \task task takes as input a \emph{premise}, multimodal information such as a sentence, audio clip, or video, and a \emph{hypothesis}, a declarative statement that may or may not be supported by the premise. The model produces a scalar \emph{prediction} in [0, 1] representing the probability that the hypothesis is true given only the presented evidence, rather than its real-world truth (\autoref{fig:method}). A claim plausible in general but unsupported by the premise should thus receive low probability. 

\paragraph{Data Collection}
Since \citet{chen-etal-2020-uncertain} provide scalar probability annotations for text, we focus on annotating audio, visual, and audio-visual premises. We build on \wikivideo \cite{martin2025wikivideo}, which provides binary support judgments for claims (serving as hypotheses) against video content (serving as premises) in the \multivent test collection \cite{kriz2025multivent20massivemultilingual}. We select 10 topics, following \citet{martin2025mirage}, for our test set and for each video-claim pair, we elicit scalar probability judgments from expert annotators. Annotators also indicate which modalities (audio, video, or both) informed their judgment. Annotators always view the full audiovisual video, but label a judgment "audio only" (or "video only") only when it could be reached from that modality alone. We provide details in \autoref{append:annotation}.



%% file: sections/40-method.tex
Our method consists of three components: (1)~\emph{self-consistent teacher calibration}, where we aggregate multiple reasoning traces from a teacher model to produce calibrated training labels; (2)~\emph{latent distribution confidence modeling}, where we probe a model's internal latent distribution; and (3)~\emph{modality-specific batching}, which groups training samples by modality to improve gradient stability and reduce padding overhead.


\subsection{Self-Consistent Teacher Calibration}
To address the lack of fine-grained probability judgments in the \clotho \cite{drossos2020clotho} and \wikivideo datasets, we employ a teacher model to generate labels. However, teacher models are poorly calibrated in zero-shot. Following prior work in self-consistency \cite{wang2023selfconsistency, xiong2024can}, 
we calibrate the teacher without training: for each premise-hypothesis pair, the teacher produces five independent responses, aggregated by either mean (average confidence) or max (majority confidence). Both strategies substantially improve accuracy and reduce MSE against human judgments, with 5-mean performing best (MSE $3.67 \times 10^{-2} \rightarrow 1.56 \times 10^{-5} $ for Qwen3-VL Thinking). 
Further details are in \autoref{append:synth_model}.

\input{tables/self_consistency}
\subsection{Latent Distribution Confidence}
To produce calibrated probability estimates, we design a model that predicts a latent distribution over confidence levels rather than generating a scalar value as discrete tokens. Inspired by \citet{wang2025always} and \citet{jiang2024addressing}, we discretize the interval $[0, 1]$ into $N$ bins $\{b_0, b_1, \ldots, b_{N-1}\}$ and associate each bin with a dedicated confidence token \texttt{<CONF\_i>}. During training, the model learns to predict a distribution over these tokens, supervised by a target Gaussian distribution $Q \sim \mathcal{N}(y, \sigma^2)$, where $y$ is the ground-truth probability label and $\sigma$ is set to a small value to approximate a sharply peaked distribution. We train this model by minimizing the Kullback--Leibler divergence~\cite{KL} between the target distribution $Q$ and the predicted distribution $P$: $\mathcal{L}_d = D_{\mathrm{KL}}(Q \| P)$. In the experiment, we set $N$ to 100 and $\sigma$ to $0.05$ (sensitivity analysis in \autoref{app:additional results}).

At inference time, a smooth probability estimate is reconstructed via weighted aggregation:
\begin{equation}
    \hat{p} = \sum_i f(b_i) \, p(\texttt{<CONF\_i>}),
\end{equation}
where $f(\cdot)$ maps each bin $b_i$ to its representative probability value. With this, we are able to capture the model's internal confidence distribution enabling smoother estimates than discrete token outputs or logit sampling. 




We compare against a \textbf{token-based baseline} that directly generates probability values as natural language tokens, trained with a standard autoregressive objective. Since outputs are produced autoregressively, the resulting probabilities are inherently discrete, limiting fine-grained confidence.

\subsection{Training}


To optimize training efficiency and gradient stability, we group samples of the same modality into uniform batches, aligning with the balance-aware strategy in \citet{li2025eagle}. Mixing modalities within a batch causes unbalanced gradient magnitudes and, due to length disparity ($L_{\text{text}} \ll L_{\text{video}}$), excessive padding that dilutes the gradient signal~\cite{dao2022flashattention,chen2018gradnorm}; modality-specific batching addresses both. We provide training dynamics in \autoref{append:training}.

%% file: sections/50-results.tex
\input{tables/binary_results}

\paragraph{Training}
We initialize our models from Qwen2.5-Omni-3B \cite{xu2025qwen25omnitechnicalreport} and report specific training parameters in \autoref{append:training}.

\paragraph{Evaluation Setup} 
We evaluate on three datasets: \clotho \cite[binary audio,][]{drossos2020clotho, deshmukh2024audioentailmentassessingdeductive}, \unli \cite[scalar text,][]{chen-etal-2020-uncertain}, and \wikivideo claims for binary and scalar judgments in audio, visual, and audiovisual. Baselines span audio-only (AF \cite{goel2025audioflamingo3advancing}, Q2-A \cite{chu2024qwen2audiotechnicalreport}), language-only (Q3 \cite{yang2025qwen3technicalreport}), vision-language (Q2.5-VL \cite{bai2025qwen25vltechnicalreport}, Q3-VL \cite{bai2025qwen3vltechnicalreport}), and omnimodal (Q2.5-O \cite{xu2025qwen25omnitechnicalreport}) models. We report accuracy for binary and MSE for scalar evaluation. Baselines generate yes/no or scalar values via an optimized prompt (\autoref{append:prompts}); scalar models use a 0.5 threshold for binary decisions. We also have an analysis for threshold choice in \autoref{append:threshold}.

\input{tables/unli_results}
\paragraph{\task-Binary}
In \autoref{tab:binary_nli}, we report binary accuracy on \clotho and \wikivideo. On \clotho, \method substantially outperforms audio baselines, improving from near chance to near perfect performance. On \wikivideo, \method outperforms all baselines across every modality and combination, suggesting that by training on a diverse mixture of modalities the model learns to ground critical evidence that larger baselines fail to align. \method{} also transfers beyond the Qwen family: trained on a Gemma-4-E2B backbone, it outperforms the zero-shot Gemma4-12B at less than half the parameters (Appendix~\ref{app:gemma}).


\paragraph{\task}
In \autoref{tab:multi_unli}, we report MSE for scalar judgments on \unli and \wikivideo. On \unli, our distribution-based model achieves the lowest MSE, outperforming text-only models up to 32B parameters. Across all \wikivideo settings, both 3B models outperform the omnimodal and modality-specific baselines, with the largest gains in the vision-only and audiovisual settings. These results demonstrate that \method is well-calibrated across modalities. The gain from distribution-based training indicates that supervising with a target Gaussian distribution helps the model internalize uncertainty more effectively, mitigating the common issue of overconfidence by providing a more granular signal for fine-grained probability estimation.


\paragraph{Token vs Distribution}
Comparing the two model variants, token-based (T) and distribution-based~(D), \method-D generally achieves lower and more stable MSE, while \method-T performs better on audio. We attribute this to the distribution-based model's ability to express smooth probability estimates via a latent distribution over confidence tokens, whereas the token-based variant is constrained to discrete outputs by the autoregressive generation process. The latent distribution approach captures internal confidence more accurately, enabling finer resolution and better alignment with human uncertainty.

\paragraph{Modality Specific}
In \autoref{tab:modality-specific}, we compare training on single modalities against training on all modalities. Single modailty models specialize in their own modality: each modality performs good on their own and improves text performance but degrades on other non-text modalities. \method achieves lowest MSE across all the \wikivideo data and remains competitive on UNLI. This reciprocal benefit suggests that signals from auxiliary modalities help the model internalize uncertainty patterns and help each individual modality.

\input{tables/modality-specific}
\paragraph{Ablations} 
We ablate each component with the same backbone, data, and LoRA configuration in \autoref{tab:ablation}. Replacing our KL objective with direct scalar regression collapses performance (49.8 on CLOTHO), confirming gains stem from the distributional objective rather than fine-tuning alone. Single-vote teacher labels and modality-mixed batching both degrade all settings, most severely audiovisual (70.1 $\rightarrow$ 56.1 and 59.8). Removing the premise (claim-only) drops WV-V from 74.6 to 46.5, confirming reliance on the provided evidence.

%% file: tables/binary_results.tex

\begin{table}[]
    \centering
    \setlength{\tabcolsep}{3.5pt}
    \begin{tabular}{lc|cccc}
    \toprule
        \textbf{Model} & \textbf{P} & \textbf{Clot} & \textbf{WV-V} & \textbf{WV-A} & \textbf{WV-AV} \\
    \midrule
        AF & 7B & 48.9 & $-$ & 18.7 & $-$ \\
        Q2-A & 7B & 48.7 & $-$ & 34.7 & $-$ \\
    \midrule
        Q2.5-VL & 32B & $-$ & 22.1 & $-$ & $-$ \\
        Q3-VL & 32B & $-$ & \textbf{81.1}& $-$ & $-$ \\
    \midrule
        Q2.5-O & 7B & 50.1 & 50.5 & 66.6 & 45.2 \\
        Gemma4 & 12B & 74.3 & 63.7 & 66.8 & 54.0 \\
    \midrule
        \method-T & 3B & \textbf{97.5} & 56.3 & \textbf{71.5} & 54.3 \\
        \method-D & 3B & \underline{95.8} & \underline{74.6} & \underline{70.2} & \textbf{70.1}\\        
    \bottomrule
    \end{tabular}
    \caption{Accuracy on \task-Binary. P: Parameters. AF: Audio Flamingo. Q: Qwen, A-Audio VL-Vision Language, O-Omni. Clot: Clotho, WV: \wikivideo, V-Vision only, A-Audio only, AV-Audio-Visual.}
    \label{tab:binary_nli}
    \vspace{-1em}
\end{table}

%% file: tables/unli_results.tex
\begin{table}[]
    \centering
    \setlength{\tabcolsep}{3.3pt}
    \begin{tabular}{lc|cccc}
    \toprule
        \textbf{Model} & \textbf{P} & \textbf{UNLI} & \textbf{WV-V} & \textbf{WV-A} & \textbf{WV-AV} \\
    \midrule
        AF & 7B & 27.2& $-$& 49.9 & $-$\\
        Q2-A & 7B & 77.8& $-$& 62.6& $-$\\
    \midrule 
        Q3 & 32B & 10.6 & $-$ & $-$ & $-$\\
        TMTO & 14B & \underline{7.5} & $-$ & $-$ & $-$ \\
    \midrule
        Q2.5-VL & 32B & 12.6 & 13.3& $-$ & $-$\\
        Q3-VL & 32B & \underline{7.5} & 10.5 & $-$ & $-$\\
    \midrule
        Q2.5-O & 7B &11.4 &  14.1& 14.3& 15.2\\
    \midrule
        \method-T & 3B & 8.1 & \underline{10.1} & \textbf{3.4} & \underline{9.8} \\
        \method-D & 3B & \textbf{5.7} & \textbf{7.8} & \underline{8.4} & \textbf{7.9} \\
    \bottomrule
    \end{tabular}
    \caption{Mean squared error (x100) for \task judgments. P: Parameters. AF: Audio Flamingo. Q: Qwen, A-Audio VL-Vision Language, O-Omni.  WV: \wikivideo, V-Vision only, A-Audio only, AV-Audio-Visual. TMTO: model from \citet{wang2025always}.}
    \label{tab:multi_unli}
    \vspace{-1em}
\end{table}


%% file: tables/modality-specific.tex
\begin{table}[]
    \centering
    \begin{tabular}{l|cccc}
    \toprule
        \textbf{Modality} & \textbf{UNLI} & \textbf{WV-V} & \textbf{WV-A} & \textbf{WV-AV} \\
        \midrule
            Text & \textbf{5.0} & 16.2 & 22.4 &  15.9\\
            Audio & 8.1 & 22.4 & 37.0 & 22.2 \\
            Video & 10.5 & \underline{13.0} & 16.3 & \underline{13.1} \\
            \method & \underline{5.7} & \textbf{7.8} & \textbf{8.4} &  \textbf{7.9} \\
    \bottomrule
    \end{tabular}
    \caption{MSE of methods trained on different modalities. Modality: Modality trained on. \method trained on all.}
    \label{tab:modality-specific}
    \vspace{-1.5em}
\end{table}

%% file: sections/70-conclusion.tex
We introduce \task, the task of calibrated probabilistic inference across text, audio, video, and their combinations. We propose \method, which combines self-consistent teacher calibration, latent distribution confidence modeling, and modality-batched training. Our results suggest that latent distribution modeling better internalizes confidence, while modality-specific batching ensures more stable training signals across diverse data. Notably, we find that training on a mixture of modalities provides a reciprocal benefit, where signals from auxiliary modalities help improve performance within any single modality. Our experiments show that our 3B-parameter model achieves lower calibration error and higher accuracy than zero-shot baselines up to 32B parameters.


\section*{Limitations}
Our evaluation set, while carefully annotated, is limited in scale: we collect scalar probability judgments on 10 topics from 4 annotators. Additionally, our training data relies entirely on teacher-generated labels rather than human-annotated scalar probabilities, as collecting fine-grained probability judgments at training scale is prohibitively expensive. While we show that self-consistent teacher calibration produces labels well-aligned with human judgment, the quality ceiling of our trained models is ultimately bounded by the fidelity of these synthetic labels. Future work could explore active learning or annotation-efficient strategies to incorporate human scalar judgments directly into training. 

%% file: appendix/related.tex
\subsection{Uncertainty Quantification and Calibration of Language Models}

Reliable uncertainty estimation is essential for the trustworthy deployment of multimodal language models.
Recent work has demonstrated that well-aligned confidence signals enable more efficient allocation of resources, either through deferring to human experts in high-risk settings \cite{jurayj-etal-2025-final} or to define early-stopping policies to minimize inference costs \cite{wang2026conformal}. These approaches differ in where and how uncertainty is measured. At the token level, \citet{skow2026rankvideoreasoningrerankingtexttovideo} show that logit-based calibration can be used to rerank videos by relevance, while ensemble methods like self-consistency \cite{wang2023selfconsistency} or LoRA ensembles \cite{wang2023lora} offer lightweight adaptations of deep ensembles \cite{lakshminarayanan2017simple} to help mitigate language models' tendency towards overconfidence \cite{xiong2024can}, and reasoning can help surface these latent signals into verbalized uncertainty scores \cite{damani2025beyond}. Research into multimodal uncertainty quantification remains nascent and struggle with robustness \cite{wang2026are}, focusing on images and relying on costly inference-time policies \cite{chen-etal-2025-unveiling}, implementing popular methods in neural network \cite{guo2017calibration} and language modeling \cite{kadavath2022language}. In contrast, we show how multimodal models can directly learn to convey their uncertainty when processing dynamic video inputs.

\subsection{Fine Grained Probability Estimation}

Conditional probability estimation \cite{chen-etal-2020-uncertain} extends traditional natural language inference \cite{bowman2015large} to measure language models' ability to evaluate the likelihood of a hypothesis given a premise on a continuous scale. This setting extends conventional binary judgments toward metrics of credence that more closely align with human uncertainty \cite{pavlick2019inherent, nie2020can}. Although language models naturally struggle to process ambiguities evident to humans \cite{jurayj2022garden, stengel-eskin2024zero}, targeted interventions such as synthetic data supervision can help mitigate this pathology \cite{wang2025always}. Although entailment \cite{xie2019visualentailmentnoveltask} and uncertainty quantification \cite{upadhyay2023probvlm} have been independently studied for multimodal problems, this line of work has been conducted almost exclusively in the text and image domain. Our work extends the uncertain inference to video-language models, showing how models learn to integrate visual, auditory, and linguistic cues into well-calibrated probabilistic judgments.

\subsection{Output Parameterization for Probability Estimation}

The choice of output parameterization shapes regression and probability estimation in ways that are largely orthogonal to task formulation. Binned distributional losses have been shown to regularize the loss landscape and reduce gradient variance relative to scalar regression objectives \cite{imani2018improving}, with subsequent analysis attributing these gains to improved internal representations rather than merely to the loss form itself \cite{imani2024investigating}. Distributional outputs also admit principled post-hoc calibration in ways that discrete or pointwise predictions do not \cite{guo2017calibration}, and latent bin-based heads yield finer-grained probability estimates than autoregressive token decoding, which is constrained to coarse decimal vocabularies (e.g., 0.5, 0.8) \cite{wang2025always}. Our findings are consistent with this literature: modeling probability as a distribution over latent bins, rather than as a generated numeric token, produces both better-calibrated and more fine-grained credence estimates in the video-language setting.

%% file: appendix/annotation.tex
\input{figures/human_agreement}

\subsection{Data Selection}
We select the same events from \wikivideo as \citet{martin2025mirage}. We do this to ensure a broad coverage of event types, dates (ensuring some are out of model parametric knowledge), and to allow for our method and others to be evaluated in applications like multimodal RAG evaluation. The full event list is: 
\begin{enumerate}
    \item Launch and commissioning of the James Webb Space Telescope
    \item 2018 lower Puna eruption
    \item Notre-Dame fire
    \item 2022 United States Senate election in Georgia
    \item Hurricane Irma
    \item 2018 Anchorage earthquake
    \item 2025 Canadian federal election
    \item 2025 Myanmar earthquake
    \item Blue Ghost Mission 1
    \item Liberation Day Tariffs
\end{enumerate}

\subsection{Annotation Protocol}
In \autoref{fig:annotation_instructions}, we provide the annotation instructions given to the annotators. Annotators are given a video and a set of claims and asked to estimate the probability (0--100\%) that each claim is true given the video. For each claim, annotators also indicate which modalities (audio, video, or both) informed their judgment. To ensure consistency, annotators are provided with a background story for each event and instructed to treat the event as currently occurring. Annotators may use general background knowledge but are prohibited from searching for information that would directly confirm or deny a claim. The annotation interface can be seen in \autoref{fig:annotation_protocol} and \autoref{fig:annotation_protocol_context}.



\subsection{Annotation Consistency and Quality Control}
The final dataset was annotated by four expert annotators. To ensure high-quality data, we first had these annotators redundantly annotate a pilot study to evaluate and ensure inter-annotator agreement. In this pilot, we found that the annotators have a Krippendorff's $\alpha$ \cite{krippendorff2011computing} of 0.7121, which indicates high agreement between the annotators. In \autoref{fig:human_score}, we also visualize other alignment calculations between these annotators with Pearson correlation \cite{pearson1895vii} and Spearman's rank correlation \cite{spearman1961proof}, Mean Squared Error \cite{bishop2006pattern}, and Jensen-Shannon distance \cite{lin2002divergence}. 



\subsection{Data Statistics} 
After verifying the consistency of the annotations, we compiled a final evaluation dataset consisting of 2,526 human-labeled video-claim pairs.

%% file: figures/human_agreement.tex
\begin{figure*}
    \centering
    \includegraphics[width=1\linewidth]{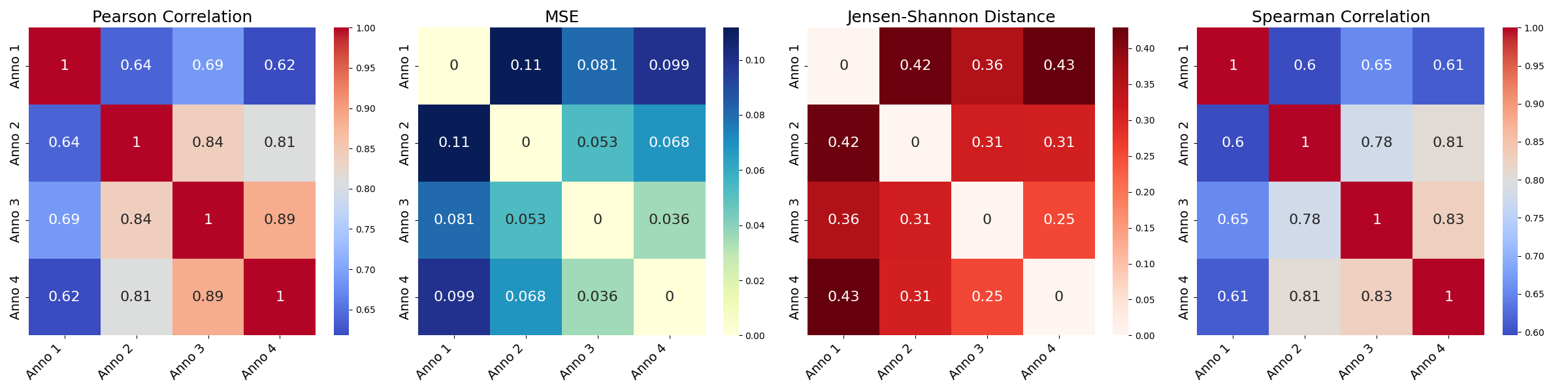}
    \caption{Annotators agreement analysis across four metrics. The heatmaps display pairwise consistency among annotators using Pearson Correlation, Mean Squared Error (MSE), Jensen-Shannon Distance, and Spearman Correlation, illustrating the reliability and alignment of human evaluations.}
    \label{fig:human_score}
\end{figure*}

%% file: appendix/synth_model.tex

We select a set of potential teacher models: Qwen3-VL-32B (Thinking and non-Thinking) and Qwen3-Omni-30B (Thinking and non-Thinking). To evaluate teacher reliability, we have each candidate operate on the same set of redundant annotations used for annotator agreement analysis. For every premise-hypothesis pair, the teacher is prompted with the same zero-shot scalar prompts shown in \autoref{prompt:scalar_audio}--\autoref{prompt:scalar_text}, with the modality of the prompt matched to the modality of the premise. Each teacher produces $N = 5$ independent responses per pair, sampled with temperature $T = 1.0$ to encourage diverse reasoning traces. From each response, we extract the scalar probability inside the \texttt{<answer>} tags.

We then aggregate the $N$ scalar predictions into a single calibrated label using two strategies: \textbf{5-mean} averages the scalar probabilities directly, capturing the teacher's expected confidence, while \textbf{5-max} applies majority voting on the binarized predictions (using a 0.5 threshold) and returns the corresponding mean probability of the majority class, capturing the teacher's modal judgment. We compare both against single-sample inference (\textbf{1}) in \autoref{tab:model_results}.

The results across three metrics(Accuracy, MSE, and Krippendorff's $\alpha$) indicate that Thinking models significantly outperform their non-Thinking counterparts under all aggregation strategies. Both Qwen3-VL Thinking and Qwen3-Omni Thinking achieve substantially higher Accuracy and lower MSE, and the gain from aggregation (5-mean and 5-max over single inference) is consistently larger for Thinking models, indicating that their reasoning traces are more diverse and complementary. The Thinking versions also show stronger inter-rater reliability with human annotators, as reflected by higher Krippendorff's $\alpha$ values. Among the two aggregation strategies, 5-mean yields the lowest MSE in nearly all cases, as averaging smooths over individual sampling noise while preserving fine-grained probability information that majority voting discards.

\begin{figure*}
    \centering
    \includegraphics[width=1\linewidth]{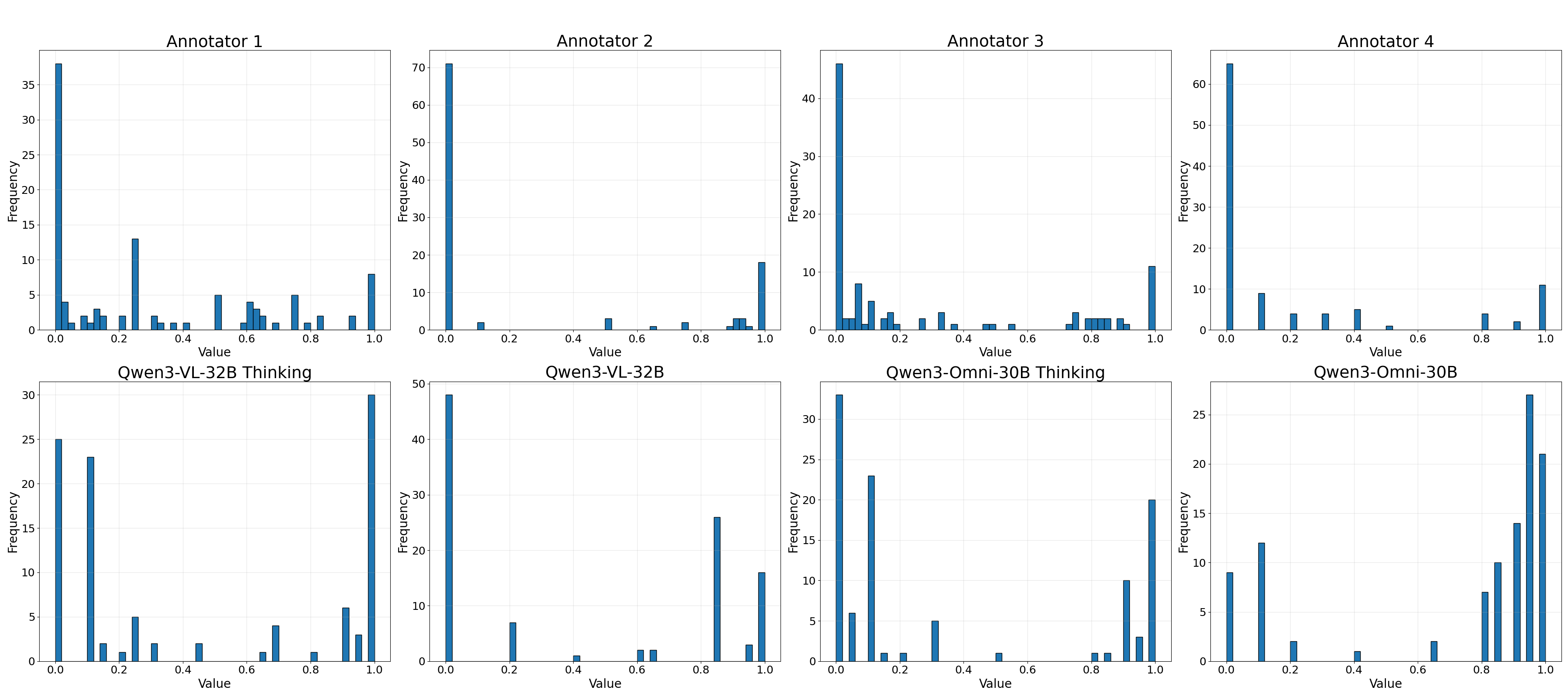}
    \caption{Probability distribution alignment between human annotators and teacher models.}
    \label{fig:synthetic_data}
\end{figure*}

In \autoref{fig:synthetic_data}, the Thinking models additionally exhibit a sharp bimodal distribution that closely mirrors the ``0-or-1'' scoring pattern of human annotators. Non-Thinking models, by contrast, concentrate probability mass in the middle range, reflecting systematic under-confidence and hedging behavior. Based on these findings, we adopt Qwen3-VL-32B Thinking and Qwen3-Omni-30B Thinking with 5-mean aggregation as our teachers for vision and audio modalities respectively, using their averaged scalar outputs as soft labels for \clotho and \wikivideo training data.

\begin{table*}[]
    \centering
    \begin{tabular}{cc|ccc}
    \toprule
        \textbf{Model} & \textbf{Votes} & \textbf{Acc.} & \textbf{MSE} & \textbf{Krippendorff's $\alpha$}\\
    \midrule
        \multirow{3}{*}{Q3-VL}
            & 1 & 0.6190 & 2.722e-2 & 0.5659\\
            & 5-max & 0.6666 & 1.391e-2 & 0.5998\\
            & 5-mean & 0.6666 & 1.296e-2 &0.5984\\
    \midrule
        \multirow{3}{*}{Q3-VL Thinking}
            & 1 & 0.6571 & 3.673e-2 & 0.5893\\
            & 5-max & 0.8571 & 8.748e-4 &0.6078\\
            & 5-mean & 0.8571 & 1.560e-5 &0.6708\\
    \midrule
        \multirow{3}{*}{Q3-O}
            & 1 & 0.3142 & 2.153e-1 & 0.4161\\
            & 5-max & 0.5333 & 6.801e-5 &0.5790\\
            & 5-mean & 0.5714 & 6.727e-2 &0.5677\\
    \midrule
        \multirow{3}{*}{Q3-O Thinking}
            & 1 & 0.7333 & 1.004e-2 & 0.6136\\
            & 5-max & 0.7619 & 5.863e-5 &0.6125\\
            & 5-mean & 0.7619 & 2.404e-5 &0.6501\\
    \bottomrule
    \end{tabular}
    \caption{Accuracy, Mean squared error and Krippendorff's $\alpha$ for teacher model evaluation. Q: Qwen, VL-Vision Language, O-Omni.}

    \label{tab:model_results}
\end{table*}

%% file: appendix/training.tex
\begin{table}[]
\centering

\begin{tabular}{lc}
\hline
\textbf{Hyperparameter} & \textbf{Value} \\ \hline
Number of Epochs & 15 \\
Learning Rate & $1 \times 10^{-4}$ \\
Warmup Steps & 5,000 \\
Gradient Accumulation Steps & 4 \\
LoRA Rank ($r$) & 16 \\
LoRA Alpha ($\alpha$) & 32 \\
LoRA Dropout & 0.1 \\ \hline
\end{tabular}
\caption{Training Hyperparameters and LoRA Configuration}
\label{tab:hyperparameters}
\end{table}

We utilize three primary datasets for training: Clotho, UNLI, and WikiVideo. Since the UNLI dataset provides fine-grained probability distributions while Clotho and WikiVideo do not, we generated synthetic soft labels for the latter two datasets to maintain label consistency across the entire training set. 

The data partitioning is as follows:
\begin{enumerate}
    \item Clotho: Development split for training; evaluation split for testing.
    \item WikiVideo: 10 events reserved for testing; the remainder for training.
    \item UNLI: Standard train/validation splits for training and testing.
\end{enumerate}

The model was trained on four NVIDIA A100 (80GB) GPUs using the DeepSpeed ZeRO-2 optimization stage. We utilized Low-Rank Adaptation (LoRA) for parameter-efficient fine-tuning \cite{hu2022lora}. The detailed hyperparameter configurations are summarized in \autoref{tab:hyperparameters}.

\begin{figure*}[htbp]
    \centering
    \includegraphics[width=1\linewidth]{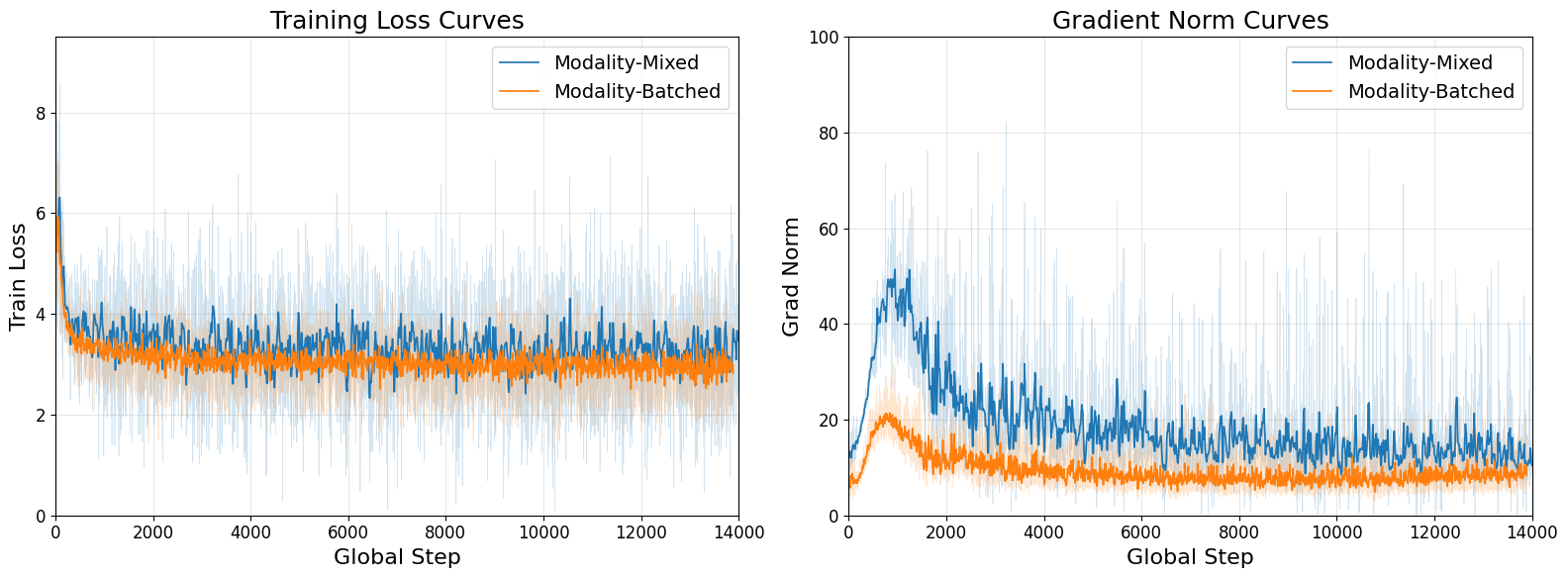}
    \caption{Comparison of Modality-Mixed and Modality-Batched strategies.}
    \label{fig:Modality Batched curves}
\end{figure*}
\subsection{Modality Batching}
To evaluate the impact of data organization in training, we compared two training strategies: Modality-Mixed and Modality-Batched. In the Modality-Mixed approach, different modalities are interleaved within each batch, whereas the Modality-Batched strategy ensures that each batch contains only a single modality.

As showed in \autoref{fig:Modality Batched curves}, the Modality-Batched strategy significantly outperforms the Modality-Mixed baseline. Specifically, the Modality-Mixed strategy exhibits substantial fluctuations in both training loss and gradient norms, ultimately converging to a higher loss value. In contrast, Modality-Batched training produces smoother curves and lower gradient norms, demonstrating that modality-level batching enhances training stability and leads to more effective model optimization.

\subsection{Model Distribution}
\begin{figure*}[!htbp]
    \centering
    \includegraphics[width=1\linewidth]{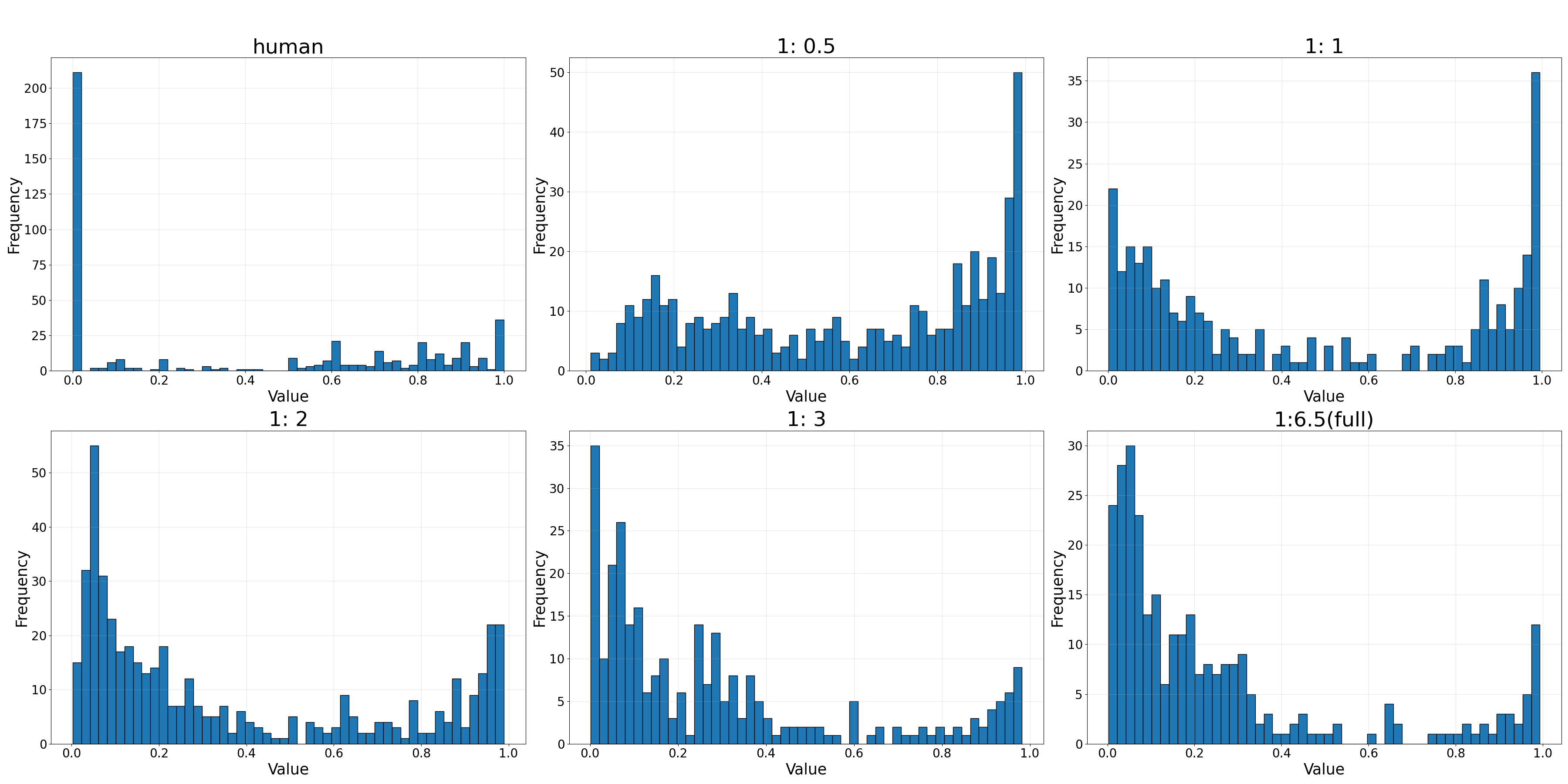}
    \caption{Probability distribution between human annotators and results from different training ratio.}
    \label{fig:ratio}
\end{figure*}

In this section, we further explore the effect of training data balance and the model's output distribution on the test set. In \autoref{fig:ratio}, we visualize the test set predictions of each ratio. Here, we see a clear shift in the distributions as the number of negative samples increases, with the expectation shifting towards 0 with every increase in negative samples. Interestingly, we see that, with the exception of 0.0-0.1, the 1:0.5 ratio seems to best reflect the human distribution. However, without enough negative signal, the model doesn't accurately discriminate negative samples.

\subsection{Modality Specific Training}

To isolate the contribution of joint multimodal training, we train three single-modality variants alongside \method and report results in \autoref{tab:modality-specific}. Each variant is initialized from Qwen2.5-Omni-3B and trained on only one of the three datasets: \unli for the text-only model, \clotho for the audio-only model, and \wikivideo for the video-only model. All other components are held fixed across runs, including the latent distribution confidence objective, self-consistent teacher labels, and the hyperparameters listed in \autoref{tab:hyperparameters} (epochs, learning rate, warmup, and LoRA configuration). At evaluation, every variant is tested on the full \task benchmark, including modalities unseen during training.

This design lets us attribute any performance gap to the training data composition rather than to differences in model capacity, optimization, or supervision signal. As shown in \autoref{tab:modality-specific}, single-modality models specialize narrowly and degrade on out-of-domain modalities, while \method with all modalities achieves the lowest MSE across all \wikivideo settings.

%% file: appendix/additional_results.tex
\newcommand{\ci}[2]{{\tiny\textcolor{gray}{[#1,#2]}}}

\begin{table*}[]
    \centering
    \setlength{\tabcolsep}{2pt}
    \small
    \begin{tabular}{lc|ccc|ccc|ccc|ccc}
    \toprule
        & & \multicolumn{3}{c|}{\textbf{UNLI}} & \multicolumn{3}{c|}{\textbf{WV-V}} & \multicolumn{3}{c|}{\textbf{WV-A}} & \multicolumn{3}{c}{\textbf{WV-AV}} \\
    \cmidrule(lr){3-5} \cmidrule(lr){6-8} \cmidrule(lr){9-11} \cmidrule(lr){12-14}
        \textbf{Model} & \textbf{P} & MSE & NLL & ECE & MSE & NLL & ECE & MSE & NLL & ECE & MSE & NLL & ECE \\
    \midrule
        AF & 7B & 27.2 & 8.87 & 38.5 & $-$ & $-$ & $-$ & 49.9 & 12.3 & 53.9 & $-$ & $-$ & $-$ \\[-2pt]
        & & \ci{26.0}{28.4} & \ci{8.58}{9.16} & \ci{37.3}{39.8} & & & & \ci{44.4}{55.5} & \ci{10.9}{13.6} & \ci{48.1}{59.5} & & & \\
        Q2-A & 7B & 77.8 & 8.85 & 37.9 & $-$ & $-$ & $-$ & 62.6 & 15.3 & 66.8 & $-$ & $-$ & $-$ \\[-2pt]
        & & \ci{26.1}{80.0} & \ci{8.29}{8.86} & \ci{36.7}{39.2} & & & & \ci{62.2}{72.2} & \ci{15.2}{17.5} & \ci{66.0}{76.2} & & & \\
    \midrule
        Q3 & 32B & 10.6 & 1.09 & 13.3 & $-$ & $-$ & $-$ & $-$ & $-$ & $-$ & $-$ & $-$ & $-$ \\[-2pt]
        & & \ci{10.1}{11.2} & \ci{0.99}{1.19} & \ci{12.2}{14.4} & & & & & & & & & \\
        TMTO & 14B & 7.57 & 0.56 & 14.7 & $-$ & $-$ & $-$ & $-$ & $-$ & $-$ & $-$ & $-$ & $-$ \\[-2pt]
        & & \ci{7.27}{7.87} & \ci{0.54}{0.56} & \ci{13.9}{15.5} & & & & & & & & & \\
    \midrule
        Q2.5-VL & 32B & 12.6 & 3.10 & 15.5 & 13.3 & 2.57 & 17.4 & $-$ & $-$ & $-$ & $-$ & $-$ & $-$ \\[-2pt]
        & & \ci{11.9}{13.3} & \ci{2.90}{3.29} & \ci{14.4}{16.7} & \ci{12.2}{14.6} & \ci{2.28}{2.87} & \ci{15.9}{19.1} & & & & & & \\
        Q3-VL & 32B & 7.53 & 0.67 & \textbf{2.98} & 10.5 & 2.71 & 11.5 & $-$ & $-$ & $-$ & $-$ & $-$ & $-$ \\[-2pt]
        & & \ci{7.18}{7.90} & \ci{0.62}{0.70} & \ci{2.12}{3.94} & \ci{9.53}{11.6} & \ci{2.46}{2.97} & \ci{10.2}{13.0} & & & & & & \\
    \midrule
        Q2.5-O & 7B & 11.4 & 1.51 & 14.2 & 14.2 & 3.94 & 16.2 & 14.3 & 3.35 & 14.2 & 15.2 & 4.06 & 18.5 \\[-2pt]
        
        & & \ci{10.7}{12.1} & \ci{1.41}{1.63} & \ci{13.2}{15.3} & \ci{12.9}{15.4} & \ci{3.64}{4.27} & \ci{14.7}{17.9} & \ci{10.4}{18.2} & \ci{2.47}{4.28} & \ci{10.8}{19.2} & \ci{13.9}{16.5} & \ci{3.74}{4.38} & \ci{16.9}{20.1} \\
    \midrule
        \method-T & 3B & 8.10 & 0.96 & 9.70 & 10.0 & 0.80 & 12.1 & \textbf{3.35} & \textbf{0.23} & \textbf{7.60} & 9.78 & 0.87 & 12.1 \\[-2pt]
        
        & & \ci{7.51}{8.88} & \ci{0.86}{1.04} & \ci{8.87}{10.5} & \ci{8.63}{11.7} & \ci{0.68}{0.94} & \ci{9.99}{14.5} & \ci{2.45}{4.36} & \ci{0.19}{0.26} & \ci{6.00}{9.60} & \ci{8.32}{11.2} & \ci{0.75}{0.99} & \ci{9.09}{13.6} \\
        \method-D & 3B & \textbf{5.73} & \textbf{0.51} & 5.92 & \textbf{7.84} & \textbf{0.50} & \textbf{8.47} & 8.40 & 0.33 & 19.0 & \textbf{7.92} & \textbf{0.50} & \textbf{8.42} \\[-2pt]
        
        & & \ci{5.40}{6.06} & \ci{0.49}{0.52} & \ci{5.08}{6.74} & \ci{7.25}{8.49} & \ci{0.47}{0.51} & \ci{7.45}{9.69} & \ci{7.46}{9.38} & \ci{0.30}{0.35} & \ci{17.7}{20.4} & \ci{7.35}{8.51} & \ci{0.48}{0.52} & \ci{7.45}{9.62} \\
    \bottomrule
    \end{tabular}
    \caption{Calibration metrics for \task judgments. MSE and ECE are reported $\times 10^{-2}$; brackets show 95\% bootstrap confidence intervals(2{,}000 resamples). P: Parameters. AF: Audio Flamingo. Q: Qwen, A-Audio VL-Vision Language, O-Omni. WV: \wikivideo, V-Vision only, A-Audio only, AV-Audio-Visual. TMTO: model from \citet{wang2025always}. Lower is better for all metrics.}
    \label{tab:calibration_metrics}
\end{table*}

\begin{figure}
    \centering
    \includegraphics[width=1\linewidth]{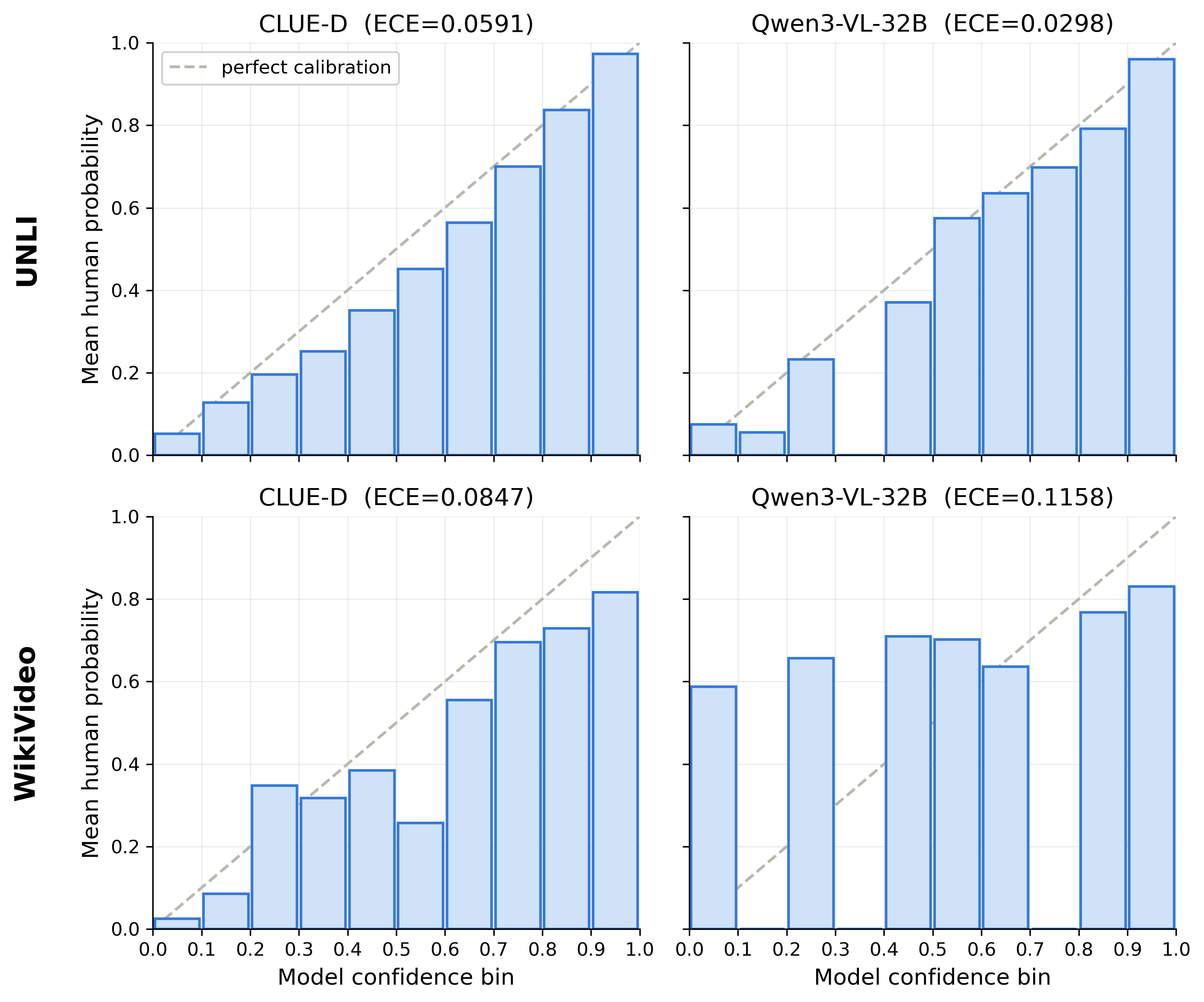}
    \caption{Reliability diagrams of \method-D and Qwen3-VL-32B on UNLI (top) and WV-V (bottom). Dashed line indicates perfect calibration; ECE reported $\times 10^{-2}$.}
    \label{fig:calibration}
\end{figure}

\begin{table}[t]
    \centering
    \setlength{\tabcolsep}{3.3pt}
    \begin{tabular}{l|cccc}
    \toprule
        \textbf{Model} & \textbf{Clot} & \textbf{WV-V} & \textbf{WV-A} & \textbf{WV-AV} \\
    \midrule
        \method-D & \textbf{95.8} & \textbf{74.6} & \textbf{70.2} & \textbf{70.1} \\
    \midrule
        w/ regression & 49.8 & 61.0 & 41.1 & 58.5 \\
        w/ single-vote & 94.6 & 72.5 & 65.5 & 56.1 \\
        w/ mod.-mixed & 95.2 & 72.5 & 63.5 & 59.8 \\
        w/ claim-only & 71.6 & 46.5 & 61.0 & 48.1 \\
    \bottomrule
    \end{tabular}
    \caption{Binary accuracy ablations of \method-D with identical backbone, training data, and LoRA configuration. w/ regression: direct scalar regression (MSE loss) replacing the KL objective; w/ single-vote: 1-vote teacher labels replacing 5-mean calibration; w/ mod.-mixed: modality-mixed batching; w/ claim-only: premise removed when evaluating. Clot: \clotho{}, WV: \wikivideo, V-Vision only, A-Audio only, AV-Audio-Visual.}
    \label{tab:ablation}
    \vspace{-1em}
\end{table}

\begin{table}[]
    \centering
    \setlength{\tabcolsep}{3.3pt}
    \begin{tabular}{l|cccc}
    \toprule
        \textbf{Config} & \textbf{Clot} & \textbf{WV-V} & \textbf{WV-A} & \textbf{WV-AV} \\
    \midrule
        $N{=}100$, $\sigma{=}0.05$ & \textbf{95.8} & \textbf{74.6} & \textbf{70.2} & \textbf{70.1} \\
    \midrule
        $N{=}10$ & 95.1 & 70.6 & 64.8 & 56.4 \\
        $N{=}200$ & 95.5 & 69.9 & 67.1 & 56.6 \\
        $\sigma{=}0.1$ & 94.8 & 70.8 & 63.7 & 56.8 \\
        $\sigma{=}0.01$ & 94.6 & 72.5 & 66.5 & 59.3 \\
    \bottomrule
    \end{tabular}
    \caption{Sensitivity to the number of confidence bins $N$ and target variance $\sigma$ (default $N{=}100$, $\sigma{=}0.05$; other components fixed). Clot: \clotho{}, WV: \wikivideo, V-Vision only, A-Audio only, AV-Audio-Visual.}
    \label{tab:sensitivity}
\end{table}

%% file: appendix/cross-family.tex
To verify that \method{} is not tied to Qwen-specific properties, we train \method{} on a Gemma-4-E2B backbone with identical training data, teacher labels, and LoRA configuration. As shown in
\autoref{tab:gemma}, \method{} (Gemma) substantially outperforms the zero-shot Gemma4-12B baseline on \clotho{}, WV-V, and WV-AV at less than half the parameters, demonstrating the framework transfers across model families.

\begin{table}[]
    \centering
    \setlength{\tabcolsep}{3.3pt}
    \begin{tabular}{lc|cccc}
    \toprule
        \textbf{Model} & \textbf{P} & \textbf{Clot} & \textbf{WV-V} & \textbf{WV-A} & \textbf{WV-AV} \\
    \midrule
        Gemma4 (zero-shot) & 12B & 74.3 & 63.7 & 66.8 & 54.0 \\
        \method{} (Gemma-4-E2B) & 5B & 94.4 & 70.5 & 53.6 & 57.5 \\
    \midrule
        \method-D (Qwen) & 3B & \textbf{95.8} & \textbf{74.6} & \textbf{70.2} & \textbf{70.1} \\
    \bottomrule
    \end{tabular}
    \caption{Binary accuracy of \method{} trained on a Gemma backbone versus
    the zero-shot Gemma baseline and the Qwen-based \method-D. Clot:
    \clotho{}, WV: \wikivideo, V-Vision only, A-Audio only, AV-Audio-Visual.}
    \label{tab:gemma}
\end{table}

%% file: appendix/threshold.tex
\begin{figure}
    \centering
    \includegraphics[width=1\linewidth]{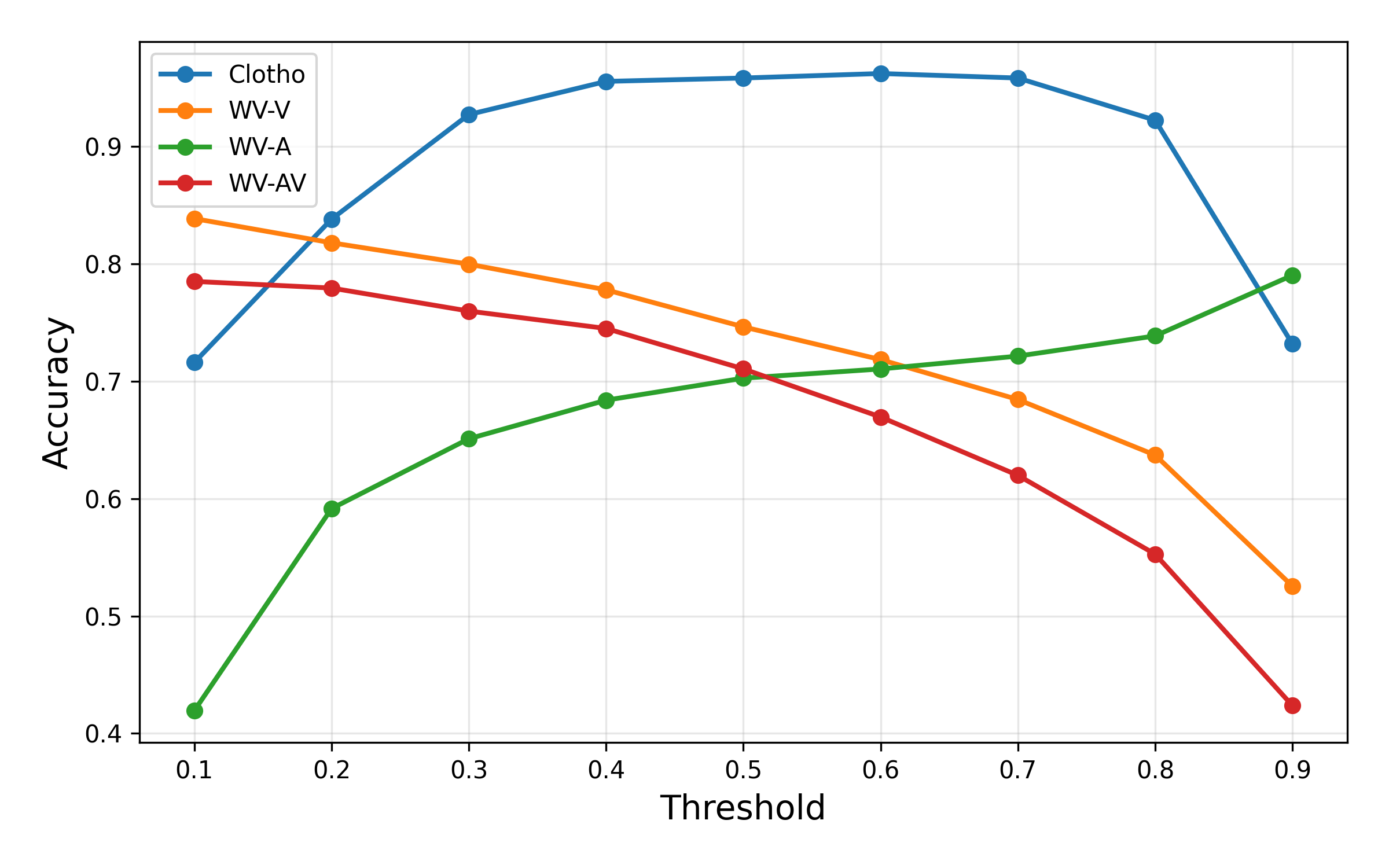}
    \caption{Binary accuracy as a function of decision threshold on \clotho and \wikivideo. WV: \wikivideo, V-Vision only, A-Audio only, AV-Audio-Visual.}
    \label{fig:threshold}
\end{figure}

In \autoref{fig:threshold}, we report binary accuracy across \clotho and \wikivideo as a function of the decision threshold. No single threshold maximizes accuracy across all settings: \clotho peaks broadly between 0.4 and 0.7, audio accuracy rises with higher thresholds, while video and omni decline monotonically. We attribute this divergence to differences in label distributions and per-modality calibration. We select 0.5 as a balanced choice, sitting near the intersection of all four curves and corresponding to the natural midpoint of the probability scale.

%% file: appendix/prompt.tex
In \autoref{prompt:scalar_audio} (audio instances), \autoref{prompt:scalar_vision} (vision instances), and \autoref{prompt:scalar_text} (text instances), we show the prompts designed to elicit scalar judgments from the models, enabling them to provide fine-grained probabilistic judgments. 

\begin{figure*}
\noindent\fbox{%
    \parbox{.98\textwidth}{%

\footnotesize
{\tt
\small
To help you make more accurate and consistent judgments, here is an expanded explanation of how to interpret and assign support percentages for audio-based evidence. These examples cover a range of real-world cases you may encounter in this annotation task.

100\% - Full and unambiguous support:  
The audio clearly and directly contains the exact event, statement, or content described in the claim. There is no need for guessing or interpretation — the claim is fully verified by the audio.

80-100\% - Almost complete support:  
The main content of the claim is clearly supported by the audio, though there may be minor ambiguity in speaker identity, context, or completeness (e.g., partial recording, mild noise). Overall, the claim is strongly supported.

60-80\% - Strong partial support:  
The audio strongly suggests that the claim is true, but some key details may be missing, unclear, or ambiguous — such as incomplete phrases, background noise, or partial conversations. The evidence is strong but not definitive.

40-60\% - Moderate partial support:  
There is some alignment between the audio and the claim, but large portions are missing, unclear, or open to interpretation. While the recording points in the same general direction as the claim, it lacks clarity or completeness for confident verification.

20-40\% - Minimal weak support:  
There are small verbal cues or contextual hints that could relate to the claim, but they are insufficient to provide confidence in its truth.

0-20\% - Very weak or speculative support:  
There may be the slightest indirect reference (such as a related topic or similar voice), but nothing concrete that verifies the claim.

0\% - No support or contradiction:  
The audio does not relate to the claim at all, or it directly contradicts it.

Based on the provided audio and text, evaluate the probability that the text statement is true.

Your answer must be a decimal number between 0 and 1, and you must strictly follow the format below:

<answer>probability\_value</answer>

Where probability\_value is the result you calculate.

The text to evaluate is:
\{text\}
}
}}

\caption{Zero-Shot scaler prompt for audio examples.}
\label{prompt:scalar_audio}
\end{figure*}

\begin{figure*}
\noindent\fbox{%
    \parbox{.98\textwidth}{%

\footnotesize
{\tt
\small
To help you make more accurate and consistent judgments, here is an expanded explanation of how to interpret and assign support percentages. These examples are designed to cover a range of real-world cases you may encounter in the annotation task.

100\% - /Full and unambiguous support:
The video clearly shows the exact event described in the claim. There is no need for guessing or interpretation. 

80-100\% - Almost complete support:
The main content in the claim is shown, but there may be minor ambiguity in location, identity, or completeness. The overall claims are supported by the video.

60-80\% - Strong partial support:
The video strongly suggests the claim is true, but some critical details may be missing, obscured, or ambiguous, limiting the ability to confirm the claim with certainty. The video gives strong but not definitive support.

40-60\% - Moderate partial support:
There is some alignment with the claim, but large portions are either missing, unclear, or open to interpretation. While the footage may point in the same general direction as the claim, it lacks the clarity or completeness needed for confident verification.

20-40\% - Minimal weak support:
There are small visual or audio cues that could hint at the claim, but they are insufficient to be confident. 

0-20\% - Very weak or speculative support:
There may be the slightest indirect reference, such as a related object or setting, but nothing concrete happens. 

0\% - No support or contradiction:
The video does not relate to the claim at all, or it directly shows something opposite. 

Based on the provided video and text, evaluate the probability that the text is true.

Your answer must be a decimal number between 0 and 1, and you must strictly follow the format below:

<answer>probability\_value</answer>

Where probability\_value is the result you calculate.

The text to evaluate is:
\{text\}
}
}}

\caption{Zero-Shot scaler prompt for vision examples.}
\label{prompt:scalar_vision}
\end{figure*}

\begin{figure*}
\noindent\fbox{%
    \parbox{.98\textwidth}{%

\footnotesize
{\tt
\small
To help you make more accurate and consistent judgments, here is an expanded explanation of how to interpret and assign support percentages based on textual evidence. These examples are designed to cover a range of logic and linguistic relationships you may encounter.

100\% - Full and unambiguous support (Entailment):
The sentence explicitly states the information in the claim, or the claim is a direct paraphrase of the sentence. There is no need for guessing; the facts are identical.

80-100\% - Almost complete support:
The main assertions in the claim are present in the sentence. There may be minor differences in wording, synonyms, or omission of non-essential details, but the core meaning is fully preserved and supported.

60-80\% - Strong partial support (Strong Implication):
The sentence strongly implies the claim is true through logical inference or context, though it may not state it explicitly. A reasonable person would conclude the claim is likely true based on the sentence.

40-60\% - Moderate partial support:
There is a topical alignment or shared keywords. The sentence discusses the same subject matter, but the specific assertion in the claim is neither confirmed nor denied. It is plausible but lacks definitive evidence in the text.

20-40\% - Minimal weak support:
There are weak textual links, such as matching entity names or a general theme, but the specific context is different. The sentence provides very little basis to deduce the claim.

0-20\% - Very weak or speculative support:
There may be a very distant connection (e.g., related vocabulary), but inferring the claim from the sentence would be highly speculative.

0\% - No support or contradiction:
The sentence is completely unrelated to the claim, or it directly contradicts the claim (proves it false).

Based on the provided sentence and claim, evaluate the probability that the claim is supported by the sentence.

Your answer must be a decimal number between 0 and 1, and you must strictly follow the format below:

<answer>probability\_value</answer>

Where probability\_value is the result you calculate.

Sentence:
\{sentence\}

Claim:
\{claim\}
}
}}

\caption{Zero-Shot scaler prompt for text examples.}
\label{prompt:scalar_text}
\end{figure*}

%% file: figures/annotation_instructions.tex
\begin{figure*}[t]
\centering
\fbox{\begin{minipage}{0.96\textwidth}
\tiny

\textbf{Fine-grained Video Annotation}

\medskip

YOU MUST READ ALL INSTRUCTIONS BELOW BEFORE COMPLETING ANY HITS

\medskip

Your goal in this task is to evaluate how likely a given claim is to be true based on a given video. You will be provided with the background story of an event, a video, and a series of claims related to the video. Your job is to determine the probability that each individual claim is true given the video, using a percentage scale.

When annotating you are allowed to use your background knowledge, but see the Background Story section for details on the type of knowledge you can use. However, for each event, you should treat it as it is currently happening (as depicted by the video) and not an event that has previously occurred. We suggest following the principle: ``you can google information to make you more confident in claims, but not that would directly confirm or deny the claim itself.''

Please follow the guidelines below when making your judgments:

100\% --- Select this if the video makes the claim almost certainly true. There should be direct and unambiguous evidence in the video that confirms the claim without the need for interpretation or inference. You should be confident in the statement beyond a reasonable doubt.

0\% --- Choose this if the video is completely unrelated to the claim or contradicts it. This means that nothing in the video provides evidence that the claim is true.

0\%--100\% --- If the video provides partial evidence for the claim, select a percentage that best reflects how likely the claim is to be true given the video. For this part, you can refer to task instructions for more details.

\medskip

\textbf{What You Will See:}

A story that provides context about the overall event. The story is to help you understand the types of background knowledge you may have, and the things you are able to search for while doing the annotation.

A video that related to an event.

One or more claims, each of which you must evaluate individually based on what is shown in the video.

\medskip

\textbf{Background Story and Knowledge}

In order to ensure that annotators make accurate and consistent judgments when evaluating the probability that a claim is true given a video, we provide a short background story for each event. The story provides contextual information that helps annotators better understand the background knowledge (and searchable knowledge) an annotator may use while annotating the task.

Without such a story, annotators may rely solely on their own assumptions or incomplete information, which can lead to inconsistent or biased annotations. By giving all annotators the same background knowledge, we aim to reduce variability in the annotation process and encourage judgments that follow a more consistent distribution. For examples of how the presence or absence of a story can affect annotation quality, please refer to the ``Examples'' section.

We suggest following the principle: ``you can google information to make you more confident in claims, but not that would directly confirm or deny the claim itself.'' To understand this better, have a look at the examples below.

\medskip

\textit{Example 1, Notre-Dame Fire}

Story: You're someone who regularly follows both local and international news. You're familiar with major cultural landmarks and global events. You are a resident of the location the event occurs in. If you are lacking background knowledge you are allowed to search for: maps of Paris and France, general knowledge a resident in France might have, etc.

Things you can google based on this event: What is a cathedral? What is gothic architecture?

Things you cannot google: How much donations were collected to restore the cathedral? When did the Notre-Dame start holding Christmas mass?

\medskip

\textit{Example 2, 2022 Senate Election in Georgia:}

Story: You are a resident living in Georgia and was going to watch a series of videos about the ongoing election. If you are lacking background knowledge you are allowed to search for: famous people, landmarks, local maps, and basic information about US elections.

Things you can google based on this event: When do elections occur in the US?

Things you cannot google based on this event: When did the 2022 senate election happen? Who won the 2022 senate election? Does Georgia have a two round election system?

\medskip

\textit{Example 3, Hurricane Irma}

Story: You are a Florida resident watching videos on news about this hurricane. If you are lacking background knowledge you are allowed to search for: maps labeling countries states, information to help you understand characteristics of hurricanes (within reason).

Things you can google based on this event: Map of Caribbean

Things you cannot google: Map of Caribbean showing the Hurricane Irma path

\medskip

\textbf{Task Instructions}

Your task involves the following steps:

1. Read the background story to understand the context.

2. Watch the entire video clip carefully. Rewatch as many times as needed to fully absorb both visual and audio details. Sometimes you can refer to suggestions in the background story to get extra information.

3. Evaluate each claim on its own, using only the evidence presented within the video itself --- including visuals, dialogue, audio cues, and on-screen text.

4. Assign a percentage score (0\%--100\%) to reflect how likely the claim is to be true given the video.

\medskip

To help you make more accurate and consistent judgments, here is an expanded explanation of how to interpret and assign probability percentages. These examples are designed to cover a range of real-world cases you may encounter in the annotation task.

100\% -- Almost certainly true: The video clearly shows the exact event described in the claim. There is no need for guessing or interpretation.

80--100\% -- Very likely true: The main content in the claim is shown, but there may be minor ambiguity in location, identity, or completeness. The video makes the claim very likely to be true.

60--80\% -- Likely true: The video strongly suggests the claim is true, but some critical details may be missing, obscured, or ambiguous, limiting the ability to confirm the claim with certainty. The video provides strong but not definitive evidence.

40--60\% -- Uncertain: There is some alignment with the claim, but large portions are either missing, unclear, or open to interpretation. While the footage may point in the same general direction as the claim, it lacks the clarity or completeness needed for confident verification.

20--40\% -- Unlikely true: There are small visual or audio cues that could hint at the claim, but they are insufficient to judge the claim as likely true.

0--20\% -- Very unlikely true: There may be the slightest indirect reference, such as a related object or setting, but nothing concrete happens.

0\% -- Almost certainly false: The video does not relate to the claim at all, or it directly shows something opposite.

These ranges are intended to help you gauge how much evidence is present and how likely the claim is to be true given that evidence. When in doubt, lean toward the lower end of the range unless you have clear reason to be confident. Remember: the key is not whether the claim is plausible in general --- only whether the video provides evidence that it is true.

\medskip

\textbf{What Does It Mean to Estimate the Probability of a Claim?}

You can think of each claim as posing a yes-or-no question: is this claim true based on the video? Your task is to estimate the probability that the answer is yes. If the video (through its visuals, audio, or on-screen text) provides evidence that the claim is true, the probability should be higher.

Evidence can come from:

Audio: Speech or sounds that indicate the claim is true.

Visual: Actions, events, or imagery shown on-screen that verify the claim.

Text: On-screen text such as subtitles, signs, or labels that confirm the claim.

If you meet something you don't understand or you are confused about, you can get some information in the story which would help. If the video does not provide any evidence relevant to the claim, or even contradicts it, then the probability should be low (near 0\%).

\end{minipage}}
\caption{Annotation instructions provided to annotators for scalar probability judgments. Along with these instructions, annotators are also provided a set of example annotations done by authors of the paper.}
\label{fig:annotation_instructions}
\end{figure*}